\documentclass{article}

\usepackage{arxiv}

\usepackage[utf8]{inputenc} 
\usepackage[T1]{fontenc}    
\usepackage{url}            
\usepackage{booktabs}       
\usepackage{amsfonts}       
\usepackage{nicefrac}       
\usepackage{microtype}      
\usepackage{lipsum}
\usepackage{graphicx}
\usepackage{subcaption}
\usepackage{csquotes}
\graphicspath{ {./images/} }
\usepackage[colorlinks=true,linkcolor=blue,citecolor=blue,urlcolor=blue]{hyperref}%

\title{Context based Text-generation using LSTM networks}

\author{
 Sivasurya Santhanam \\
 Institute for Software Technology \\
 German Aerospace Center(DLR)\\
  \texttt{sivasurya.santhanam@dlr.de} \\
}

\begin{document}
\maketitle
\begin{abstract}
Long short-term memory(LSTM) units on sequence-based models are being used in translation, question-answering systems, classification tasks due to their capability of learning long-term dependencies. In Natural language generation, LSTM networks are providing impressive results on text generation models by learning language models with grammatically stable syntaxes. But the downside is that the network does not learn about the context. The network only learns the input-output function and generates text given a set of input words irrespective of pragmatics. As the model is trained without any such context, there is no semantic consistency among the generated sentences.

The proposed model is trained to generate text for a given set of input words along with a context vector. A context vector is similar to a paragraph vector that grasps the semantic meaning(context) of the sentence. Several methods of extracting the context vectors are proposed in this work. While training a language model, in addition to the input-output sequences, context vectors are also trained along with the inputs. Due to this structure, the model learns the relation among the input words, context vector and the target word. Given a set of context terms, a well trained model will generate text around the provided context.

Based on the nature of computing context vectors, the model has been tried out with two variations (word importance and word clustering). In the word clustering method, the suitable embeddings among various domains are also explored. The results are evaluated based on the semantic closeness of the generated text to the given context.
\end{abstract}


\section{Introduction}
In the past few years, Natural language processing(NLP) has seen rapid developments and the reasons are mainly attributed to the developments in learning based algorithms and advancements in computational power. As the availability of computational resources increased, active research and development happened in the architecture of deep learning algorithms. The use of feed-forward neural networks in lots of applications yielded impressive results, yet it is not considered optimal for the language models. This is because, the feed-forward network tries to optimize the input-output function without any concern to the order of inputs. Considering the type of data used in Natural language processing (both speech and text) the sequence of encoding is important; similar to a time series data. In case of text processing, there is a significant amount of information embedded in the word order. Thus, the usage of feed-forward networks result in loss of information. In order to capture the time and sequence, recurrent neural networks are applied in the field of Natural language processing.

Recurrent neural network(RNN) is a type of sequence based neural network which has memory units to remember past information. In feed-forward networks only the output of the previous layers are fed into the next layer. In case of RNNs, the output of each cell(neuron) is given back to the same cell as input and hence the name recurrent networks. Due to this recurrent nature, the positional information of words in sequences are encoded as delays in the memory unit. This helps the recurrent network to keep track of word order. The current state of the RNN cell is calculated from the input to the current cell and the output of the previous state. Thus, in addition to the output from the previous layer, the output of the current layer at previous time stamp is also fed as input to the current layer. To demystify the time based delays in state representations, the RNN network is usually unfolded through time and seen as feed-forward network. 

\subsection{Working of RNNs}
\begin{figure}[!h]
\hspace{0.15\textwidth}
     \begin{subfigure}[b]{0.25\textwidth}
         \centering
         \raisebox{10mm}{\includegraphics[width=\textwidth]{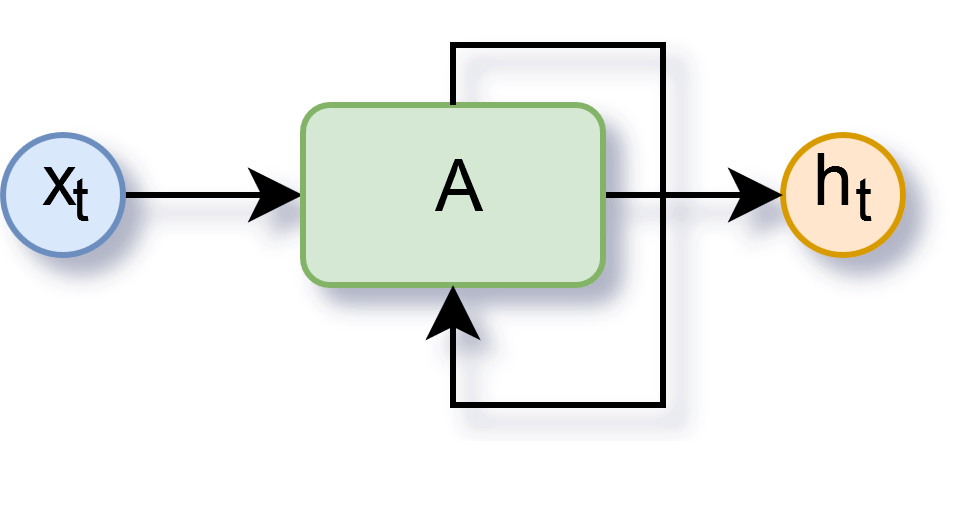}}
         \caption{RNN in state representation}
         \label{fig:rnna}
     \end{subfigure}
     \hspace{0.2\textwidth}
     \begin{subfigure}[b]{0.25\textwidth}
         \centering
         \includegraphics[width=\textwidth]{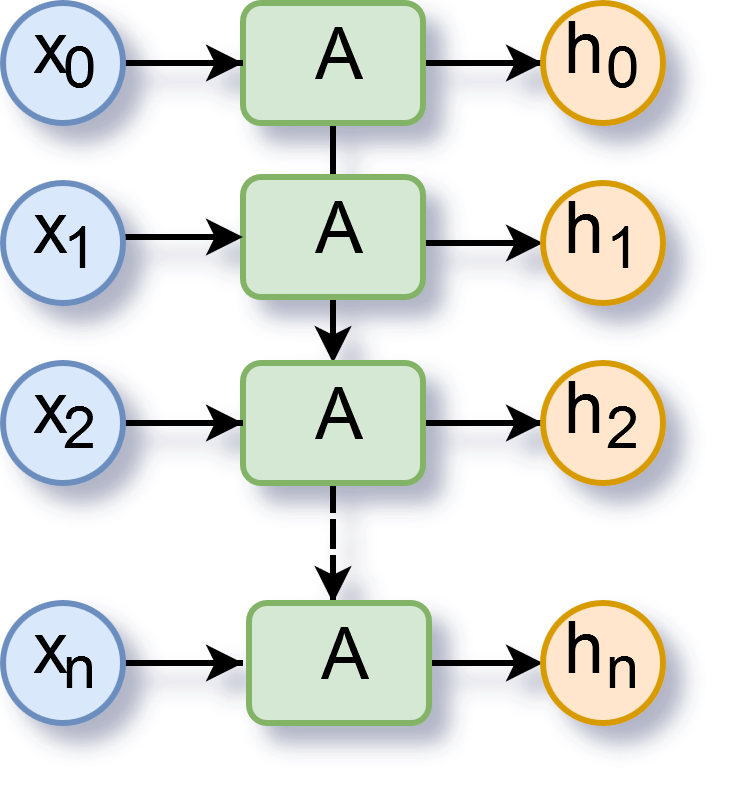}
         \caption{unfolded RNN sequence}
         \label{fig:rnnb}
     \end{subfigure}
        \caption{Architecture of RNN}
        \label{fig:rnn}
\end{figure}

Figure~\ref{fig:rnna} shows the architecture of a vanilla RNN in the state representation and Figure~\ref{fig:rnnb} shows the representation of the same RNN when unfolded based on the time information. Vanilla network is a common name for representing a generic model without any special case or variations in it. The \enquote{$x_t$} represents the input, \enquote{$h_t$} the output and the \enquote{A} block represents the hidden layer containing a single activation function. For example the sentence \emph{``The quick brown fox jumps''} is encoded as \emph{``The($X_0$), quick($X_1$), brown($X_2$), fox($X_3$),  jumps($X_4$)''}.  Similar to a feed-forward network, the model learns via back propagation. Since the inputs are given as sequences in RNNs, the back propagation also occurs with respect to the time factor and it is termed as back propagation through time (BPTT). RNNs works fine for short sequences. But in case of longer sequences, the gradients are unable to be propagated through many cells. It is referred to as the vanishing gradient problem~\cite{bengio1994learning}.

\subsection{LSTM, a variation of RNN}
One way of looking at this vanishing gradient problem is that the network is trying to propagate all the information through longer sequences. And with longer sequences, the gradients are getting diminished during BPTT. To remedy this issue, Long short term memory network (LSTM)~\cite{hochreiter1997long} is introduced. LSTMs manage to retain long-term information by propagating only the relevant information back. LSTMs are a variant of the RNNs, where instead of a single activation function in the hidden layer, there are multiple gates. The input gate, forget gate, output gate and the cell states. At each time point, the gates will determine which of the past information is relevant to retain and which has to be forgotten by throwing them away. The input gate controls the current cell information, the forget gate regulates how much amount of information from previous state has to be forgotten and the output gate filters the information to be passed on to the next layer. Due to such a nature, LSTMs could preserve history for long sentences and have been used extensively in NLP applications from question-answering systems to machine translation.

For instance, consider there are two language models(LM) which are trained on vanilla RNN and LSTM respectively. And both these models are tested on a sentence \emph{``I have participated in lots of Chess competitions since my childhood and that made me really good at playing \underline{\hspace{1cm}}''}. Since the vanilla RNN model only retains short term memory, it only considers the last phrase ``really good at playing'' and would predict words such as ``piano, football, cards'' depending upon the probability of occurrence and the dataset in which it is trained on. Whereas the LSTM model will preserve the long term information about chess competitions and would predict the word ``chess''. It is also possible that the RNN network might predict ``chess'', but the confidence score will be lower compared to an LSTM model.

Based upon the length of input and output sequences used to train the RNN/LSTM model, the configurations maybe one-to-many, many-to-many, many-to-one. Machine translation is a classical example of many-to-many configuration where the input will be a sequence in the source language and the output will also be a sequence in the target language. Sentiment analysis and intent classification comes under many-to-one where there will be an input sequence and an output class to predict. many-to-many configuration is fondly called as Sequence to sequence models or Seq2Seq

\section{Related work}
Several variations of LSTM models have been proposed by making changes in the internal gates. Gated recurrent unit(GRU) introduced by Cho et al.~\cite{cho2014learning} is a generalized simpler model of LSTM with just reset gate and update gate. Tai et al.~\cite{tai2015improved} has attempted to use tree based topology on syntactic dependencies, rather than a linear model of LSTM network.

Feeding the input sequences in the order of words lets the model learn from the past history. Schuster et al.~\cite{schuster1997bidirectional} proposed Bidirectional RNN where the input sequences are not only trained from first word to last, but also in the reversed sequence. This worked effectively for machine translation tasks~\cite{sutskever2014sequence}, as for some target languages the grammatical structure is different from the source language and training simultaneously in both directions let the model also to account for future input words.

RNNs used for language modeling based on character level encoding is done by Sutskever et al.~\cite{sutskever2011generating}. Graves~\cite{graves2013generating} demonstrated how LSTMs are used in text prediction (language modeling) and applied the same in handwriting prediction tasks. Sutskever et al.~\cite{sutskever2014sequence} proposed the general model of end-to-end approach on applying Bi-directional multi layered LSTMs for language translation.



\section{Problem statement}
It is evident from the previous works that LSTMs are widely applied in NLP tasks. From machine translation to question-answering systems, the configuration of the network is quite similar i.e, Sequence to Sequence models. These encoder-decoder seq2seq networks are good at capturing intrinsic features by mapping input to output sequences. Building a good language model which is able to capture semantic context is more than just input-output mapping function. For instance in text generation model, the network only learns the function of mapping a sequence of words to the successive word. Without necessary contextual information, the network just spits out grammatically stable but pragmatically nonsensical sentences. Thus, as the network goes on generating sentences, the sentences will have no coherence nor specific context.

Contexts provide additional semantic information about the sentences. In linguistic point of view, Dash~\cite{dash2008context} discussed how the contextual information influences the meaning of words. Gale~\cite{gale1992work} used context based methods for word sense disambiguation. Maynard et al.~\cite{maynard2000terminological} exploited the contexts in terminological acquaintance by clustering them. 

This work is based on the hypothesis that, training neural network models along with relative contextual information helps to learn about semantic association of the sentences to the context. This also helps to avoid lexical ambiguity. Based on this hypothesis, I propose a method of including contextual information to train a language model using LSTM networks. In the view of such contextual information, the model not only learns about the input-output function, but also the relation of the context words with them. Such a language model can be applied for context based text generation task, where the network generates sentences based on a given set of context words.




For a text generation problem, the performance of the model is usually evaluated between not duplicating the same text and generating new sentences which are syntactically strong. Perplexity is a good measure for it. The proposed work focuses on the degree of semantic closeness between the provided contexts and the text generated, not of the quality of generated texts. Various methods of extracting contextual information from the text is proposed. Each method is evaluated based on the semantic closeness of generated sentences for the given context using cosine similarity measure. The proposed method of training language model along with context is unanimous with any architecture of LM. Thus the performance of LSTM networks will not influence the proposed model and will not be evaluated in this work.

\section{Context based text generation}\label{sec:textgen}
Text generation is one of the applications of language models in Natural language generation(NLG) field. For the text generation task, the first step is to train a language model(LM) on a dataset of choice. For some given input seed words, the trained model predicts the succeeding word. The predicted output word is then appended with the existing input words and given as new input. This process is continuously repeated by shifting the window to generate text. The language model can be trained on word-level or character-level. In this work, I have trained the language model on word-level. Being an self-supervised learning method, there are no limitations or bias in choosing the dataset. To train the LM, a sequence from a sentence is given as input and the successive word in the sequence is assigned as the predictive word(target word). Pairs of input sequences and predictive words are generated from any dataset. While training with several instances, the network approximates a function of predicting the next word for a given set of words in sequence. For the input words \emph{i,...,i+n}, the target word is \emph{i+n+1}, where \emph{i} represents the iterator for generating training examples in the dataset and \emph{n} represents the length of the input in terms of words. Depending upon the number of network layers, dataset size, the training process may require heavy computational power. For applications like text generation, datasets which have the form of a story is suitable, as the generated text will have specific domain knowledge.

\begin{figure}[h]
  \centering
  \includegraphics[width=8cm]{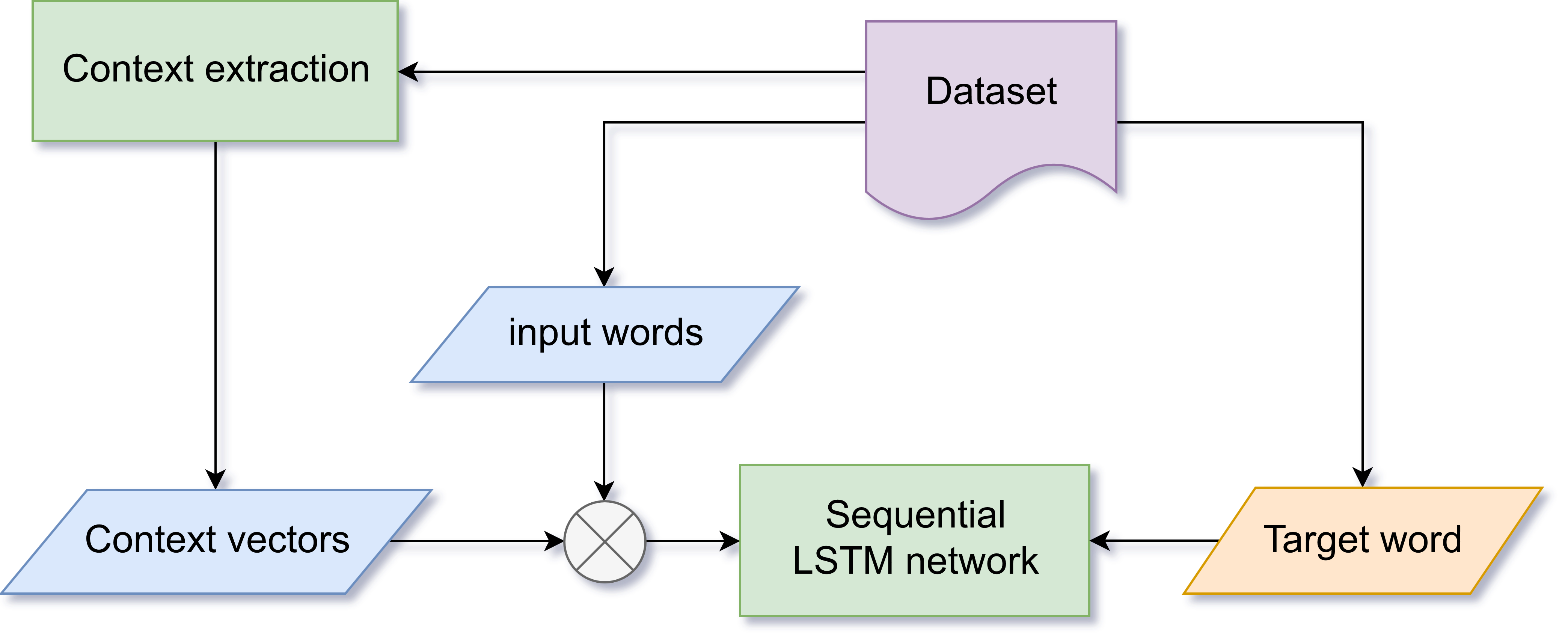}
  \caption{Proposed method architecture}
  \label{fig:architecture}
\end{figure}

The proposed model of context based text generation adopts the same strategy with a variation in the configuration to include the context along with input-output instances. As shown in the Figure~\ref{fig:architecture}, context extraction methods are applied to extract context vectors, which are appended along with input words to train the language model. 

I have proposed two different methods to extract the context vectors and both are discussed in Section~\ref{sec:methodology}. Irrespective of the method applied to obtain the context vectors, these vectors are of the same dimension as the input and output vectors. The proposed system is based on the hypothesis that the context vector of a sequence preserves the semantic information of the sequence. For instance, the context vector for the sentence \enquote{The quick brown fox jumps over the lazy dog} contains words such as \enquote{animals, forest, action}. For every input-output instances, a context vector is computed to train the context-based text generation model. A context could be just a keyword, a set of words or a topic, anything that represents the semantic content of the sentence. Context vectors are similar to the paragraph vectors~\cite{dai2015document}, but they do not reflect the complete semantics. The use of topic modeling techniques such as Latent dirichlet allocation(LDA) or latent semantic analysis(LSA) is unsuitable for our application as the nature of the topic modeling techniques are based on distribution of topics over the document. Whereas in case of text generation models, domain based dataset lacks the diversity of topics.

Of all the various configurations, the text generation uses many-to-one structure. In many-to-one, the model is trained against multiple input words and a single target word. It is worthy to note that we deliberately avoided tokenization of sentences from the dataset,  because the features such as sentence completion, sentence openings and the semantic flow from the previous to the following sentences are preserved when trained as continuous words in a paragraph. Thus the input sequences extracted from the dataset have uniform length but are not complete sentences. On the other hand, context vectors are calculated for each valid sentences. While generating training dataset, the inputs are \emph{i,...,i+n} words with output word at \emph{i+n+1} from a paragraph. The maximum number of words in the input sequence that constituted a sentence determines the context vector for that input instance. The context vectors are just appended along with the inputs when training the model. 

\section{Methodology}\label{sec:methodology}
A section of \emph{The Lord of the Rings} dataset\footnote{https://github.com/wess/iotr} is used in this work. For the data pre-processing, punctuations are removed and only sentences with a minimum of 7 words in length are used for the training. total number of training sentences are also limited to 100,000, so as to keep the computational memory space in control. As explained in Section~\ref{sec:textgen}, the input-output training instances are generated from the dataset. All the words are vectorized using one-hot encoding, thus each word is represented by a vector of dimension \emph{V}, where \emph{V} is the size of vocabulary. The length of the input sequences excluding context vector is set to 8. The context vector for a training sample is also a V-dimensional vector. The encoding process of the context vector depends upon the method of extraction. As discussed, it is a many-to-one configuration with the inputs containing the context vector appended with input words and the output contains the succeeding word in the sequence.

A sequential model of LSTM network is applied with the configuration: \emph{Bidirectional LSTM, 256 hidden units, `relu' activation} for the inner layers and \emph{`softmax` activation} for the outer layer. Keras~\cite{chollet2015keras} library is used to train the model. Since the language model is a prediction problem, that the model has to predict the one-hot encoded target word, `softmax' is suitable to determine the maximum probability of a word. Implementation is done in python with the help of libraries, including but not limited to scikit-learn, numpy and gensim. Once the model is trained along with context vectors, the prediction phase requires two sets of inputs from the user. The first input is the sequence of seed words. Seed words are the opening words given to the network to initiate the text generation process. The network expects 8 seed words, as we have trained the network with an input sequence of length 8. The second set of input is context words. One can provide any number of context words as needed. For \emph{C} context words, the network generates \emph{C} sentences. During prediction phase, the network sets the input with the first context word along with the initial seed words and predicts the subsequent word. Once the word is predicted, the first word from the seed word is removed and the predicted word is appended thus maintaining the sequence length as 8. With the new input sequence and the same context word the subsequent word is predicted. This process is continued until the sentence is completed. The sentence completion is identified by a full-stop. Once a sentence is completed, the next context word is used for generating the second sentence following the same procedure but instead of seed words, the words generated from the previous sentence is considered. The two methods of extracting context vectors are described as follows:

\subsection{Word importance method}
The word importance method is based on the hypothesis that, the most important word in the sentence represents the context of the sentence. Term frequency-inverse document frequency (tf-idf) is a classical measure of determining the importance of a word in the document/sentences. Based on the frequency of the word in the same sentence and in the entire document, the word importance(tf-idf value) is calculated. Depending upon the tf-idf value of every word in a sentence, the word having the most importance is picked out and chosen as the context word. In this method, the context is clearly a keyword picked out of the sentence. This process is executed for all the training sentences and context words are extracted. Since, the context is simply a word from the vocabulary, the context vector is represented as one-hot representation of the word. The language model is trained along with the context vectors and tested on some random context words. The results of the text generation using the current method is shown below

\begingroup
\advance \leftmargini -1em
\begin{quote}
\texttt{Context provided:} ``Hobbits – gollum – adventure – king – ring – war – friends – war – book – home''
\vspace{1mm}

\texttt{Generated text:} \it \textbf{(Hobbits)}  Hobbits lived in the woods and once an elf came all his There was Tom to go on by this If he meant all the game but there was a bit of Moria of and Bilbo was running a black eyebrows at the Mountain pretty of and and as you played up about and he said up about the hall in all this They will not come down in it. \textbf{(gollum)} He had been going up a bit and crept in into the side of a magic beard and into the prisoners of our If you have get him or I will get my dear sir and Bilbo felt he must have to find this adventures but you are cook the question they all are yes. \textbf{(adventure)} Dont me on to Bilbo my sudden understanding a pity mixed with various noticing down and their parlour as well as he could. \textbf{(king)} No trouble He was in a boat and a breakfast and their sun had been and one of his second wealth as Gandalf dared their ponies behind they had just had a fair breakfast as they had been gone further he had been adventures many He had and been very He had come a ring of their wild Well all this That
\end{quote}
\endgroup

From the result, we can conclude that the generated text does not have a good syntactic structure, but as previously said the work focuses on context words and the semantic relation to the generated words and not the performance of text prediction model. Also the generated sentences did not reflect the context from the provided keywords as expected. The proposed hypothesis did not hold. The word with highest tf-idf score was not a good representation of the context of the sentence. There was too much loss of information, as only one word is considered.

\subsection{Word clustering method}
Despite picking up the most important word as context, this method tries to find a context vector using all the words in a sentence. It is based on the hypothesis that a context vector in a high dimensional word vector space captures the semantic content of the sentence. Word embeddings are vectorized representations of words encoded with certain semantic information. Word2Vec~\cite{mikolov2013distributed}, Glove~\cite{pennington2014glove}, FastText~\cite{mikolov2018advances} are some of the commonly used word embeddings. I have chosen Word2Vec as the embedding model for the word clustering method. Word embeddings have the ability to preserve two interesting properties: semantic similarity and semantic relationships. Extracting context vectors using word clustering method works by exploiting these two features. This method has the hypothesis that, the context of a sentence is determined by a context vector present in the word vector space, such that the semantic similarity of the context vector and the accumulated sentence vector is highest. Given a list of sentences $S$ and the cluster centers $C$ of the word vector space, the hypothesis is expressed as in equation(\ref{eq:hyp})

\begin{equation}
\forall s \in S, \exists c \in C \textrm{ s.t}, \textstyle \sum_{w_i}^{w_n} \cos(s ,c) \textrm{ is minimum}
\label{eq:hyp}
\end{equation}

The accumulated sentence vector is computed by the summation of all the word vectors in the sentence. Based on the property of semantic relationships, the sum should preserve the semantic content of all the words present in it. From the property of semantic similarity, the closest vector to the accumulated word vectors is obtained from a word vector space. The word vector space is a high-dimensional space which contains all the vocabularies of a dataset represented in word embeddings, where the words with the similar semantic meaning exists closer to each other. Such word vector space is generated by training a self-supervised word2vec model over a dataset. These word embeddings(200-d vectors) constitute the word vector space.

It is clear that the language model is trained using \emph{The Lord of the Rings(LOR)} dataset, but it is possible to construct the word vector space from any dataset. I have constructed the word vector space from two sources. The first one, \enquote{domain vector space} is generated using the same LOR dataset thus having an enclosed domain of vocabulary pertaining to the story. The second one, \enquote{wiki vector space} is built from various wikipedia sources\footnote{http://wortschatz.uni-leipzig.de}~\cite{goldhahn2012building} which covers a comprehensive set of vocabulary. Experimenting with both these vector spaces helps to identify the features which best captures the context of a sentence. As a quick comparison, the context for a sentence is extracted from both vector spaces. For the sentence ``\emph{It was often said in other families that long ago one of the Took ancestors must have taken a fairy wife}'', the \emph{Domain vector space} contained context words of ``\emph{took, drew, gave, taking, unimpeachable}'', whereas the context words extracted by \emph{wiki vector space} contained ``\emph{father, mother, son, daughter, grandmother}''. It could be seen that the \emph{domain space} contained localized words of the action, whereas the \emph{wiki space} captured words of subjects and nouns in a broader context.

\begin{figure}[t]
  \centering
  \includegraphics[width=8cm]{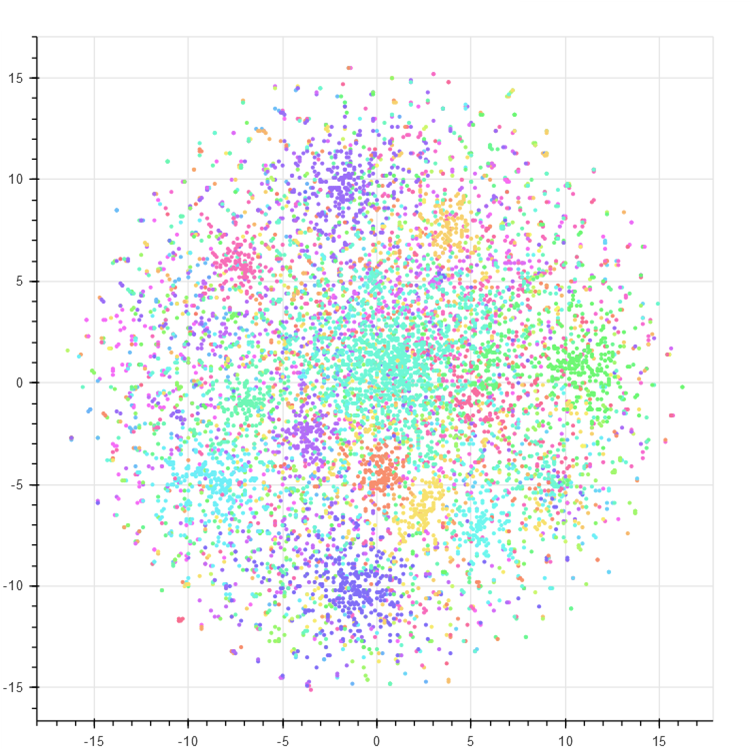}
  \caption{Word-embedding clusters(n=100) visualized under t-SNE}
  \label{fig:wordclusters}
\end{figure}

Once the word vector spaces are built, the word vectors in the high-dimensional space have been clustered using k-means clustering into multiple clusters (n=100). The clustering is done to group the semantically similar words into a cluster, thus sentences having closer meanings will obtain the same context vector. For visualizing the clusters, 200-dim vectors from the \emph{wiki space} has been resized into 60-dim vectors using Singular value decomposition(SVD), and then t-stochastic neighbor embeddings(t-SNE)~\cite{maaten2008visualizing} has been applied to reduce it down to 2 dimensions as shown in Fig~\ref{fig:wordclusters}. Obviously visualizing 100 different colors may look blurry, still the various clusters could be observed.

After clustering the vocabularies, the context vector for a sentence is obtained from the cluster center having the minimum cosine distance with the accumulated sentence vectors. Thus for each and every training instance, the corresponding cluster center is calculated. From the cluster centers, the words present in the cluster is selected as the context words and the context vector is represented with the bag-of-words(BOW) encoding. Applying all the words from the cluster for a context vector reduces the significance of the context. Thus top n words(n=5) closer to the cluster center is chosen to represent the context vector.

When the training instances are prepared from both the domain and wiki spaces, I trained the context based text generation model with the same configuration as described before. I tested the trained model on \emph{wiki vector space} and the results are shown below.

\begingroup
\advance \leftmargini -1em
\begin{quote}
\texttt{Context provided:} ``gandalf – ring – war – friends – snake – book – home – king – hobbit''
\vspace{1mm}

\texttt{Generated text:} \it
\textbf{(gandalf)} characters hobbits lived in the woods and an elf said that please try down they all that been had only been so they had only been much in the meantime in the deal of the and and they had brought the dwarves of the dwarves but they had brought him from the road and the goblins went out and again that they could hear him on their chief and the great Master of the town that of the great wooden spur which they had had been as the Great Goblin and his little sword that had was in a way but he had picked up in the dark.
\textbf{(ring)} Some ago it was so all that so fitted off he found he more one all his second staff on the floor and the laughter of the others of the door and the laughter of the line of the dwarves and he had never a shocking of the below.
\textbf{(war)} end of their way had would be and the most of the dwarves were built with guards.
\textbf{(friends)} It was like a pull for the jug.
\textbf{(snake)} smash and but the various birds that is the more and horses now he never a cave and was not to be done in the obscurest words.
\textbf{(book)} Bilbo I have a little time for the terrible hours and in fact and besides of the town were me that the track had come from a good way that was too I dont know but I am your last Burglar said a share of the rate.
\textbf{(home)} The heart of the hobbit and made of yellow wood.
\textbf{(king)} Bilbo was more and a long deal of the dwarves they had been busy in the direction of the Mountain.
\textbf{(hobbit)} The dwarves that was in a story of the chief of the dwarves and they were all alone and that they were all alone and whether they had brought of them the dwarves were not in the valley and they were forced to pull the frightful Eagle the Great Mountain and coming
\end{quote}
\endgroup

From the results, the semantic content of some of the generated sentences are in line with the context word provided. For instance, the context ``war'' generated phrases like \emph{built with guards}. ``snakes'' produced words around the forest topic like \emph{birds, horses, cave}. The context ``home'' was pretty interesting and sounded like an idiom \emph{The heart of the hobbit and made of yellow wood}. It was not successful for all the contexts. For the context ``friends'', the generated text was terrible, even for the grammaticality. The following section discusses in detail about the evaluation of these results.

\section{Results}\label{sec:results}
Before comparing the results of the methods, it is important to go through how the language model learnt over time. Changes in loss and accuracy on training and validation sets.

\begin{figure}[!h]
     \begin{subfigure}[b]{0.3\textwidth}
         \centering
         \includegraphics[scale=0.3]{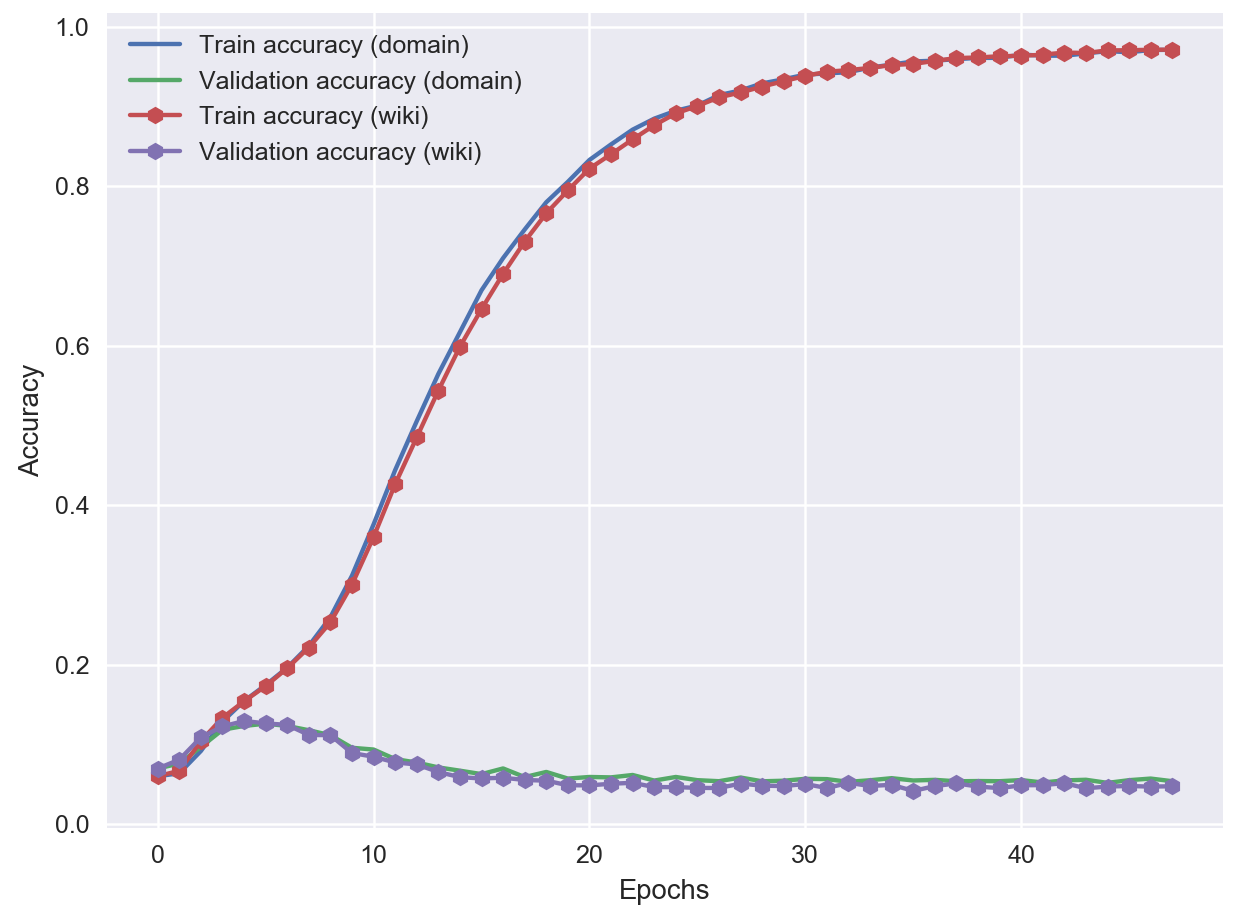}
         \caption{Training and Validation accuracy over time}
         \label{fig:accuracy}
     \end{subfigure}
     \hspace{0.2\textwidth}
     \begin{subfigure}[b]{0.3\textwidth}
         \centering
         \includegraphics[scale=0.3]{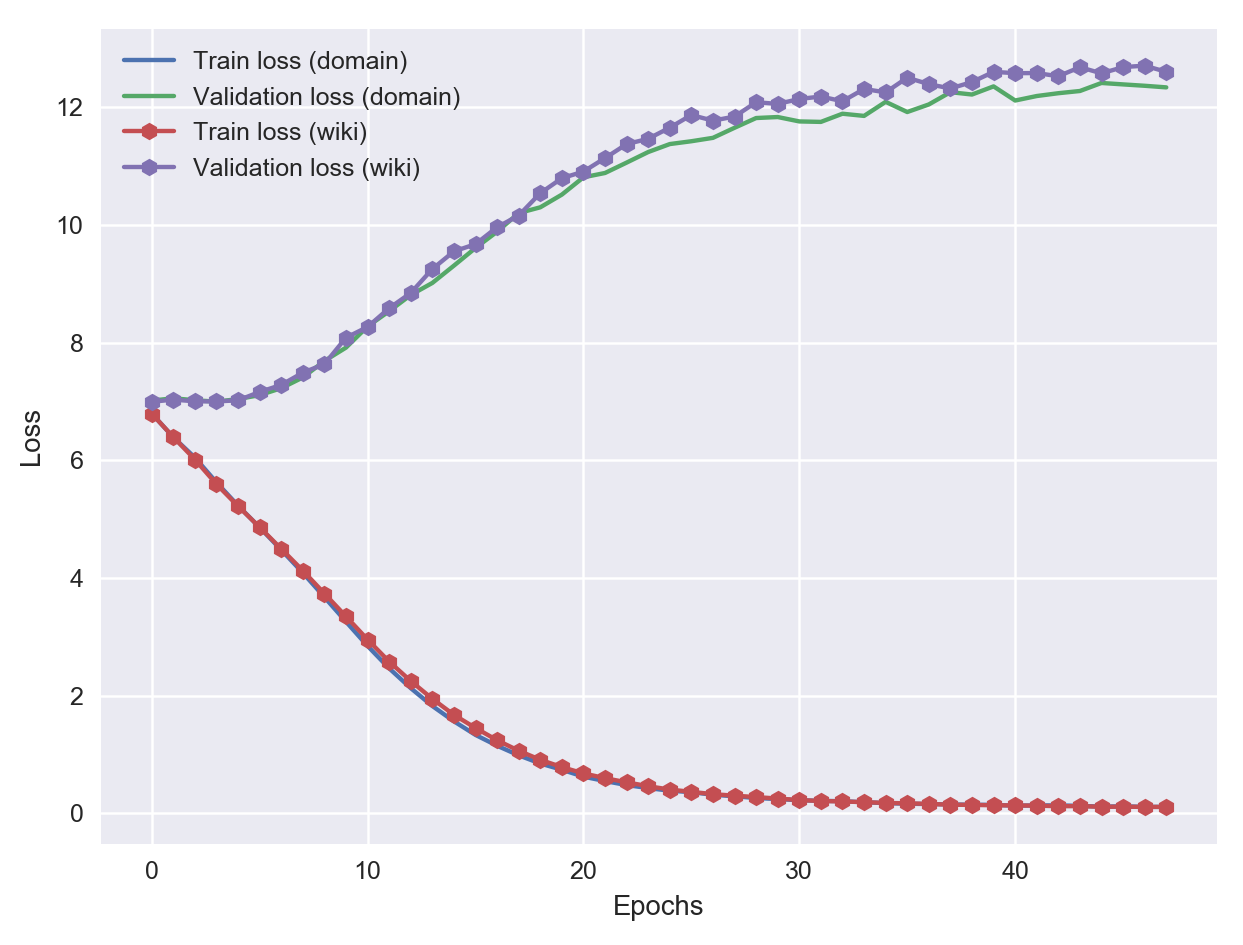}
         \caption{Training and Validation loss over time}
         \label{fig:loss}
     \end{subfigure}
        \caption{Training \& validation performance}
        \label{fig:accval}
\end{figure}

Figure~\ref{fig:accuracy} shows a plot of accuracy over epochs(time) on training and validation dataset. It compares both the word vector spaces(\emph{wiki and domain}). One could observe that the validation accuracy never really took off. The training dataset keeps on getting better and better and even reached around 97\%. It is very clear that the model is overfitting. But the problem with text generation models is that unlike other models, there is no sweet spot where the training accuracy increases and validation accuracy decreases that training the model should be halted. We are looking for a generalizable model that could generate new text, thus the whole point of validation dataset fails here.

It comes as no surprise when we also look into the loss plotted against epochs in Figure~\ref{fig:loss}. The training loss decreases while the validation loss keeps on increasing. One interesting thing to note in both the plots is that, both the \emph{wiki and domain spaces} are learning at the same rate. From these plots we will never know which vector model representation could generate better text. The only solution is to run evaluations on each and every epoch for both vector spaces to find out the best model to extract context features. In addition to comparing which model performs better, from the evaluation score at each epochs we could also ascertain the specific point of stopping for the optimum output.

There are various evaluation methods used in this area of research such as BLEU, F-measure, WER, accuracy, perplexity, etc. To evaluate against a reference data like in machine translation or question-answering downstream tasks, metrics such as BLEU, NIST, F-measure could be used. As for applications without any reference text, perplexity has been applied to find out how good is the generated language model. They are used to test the syntactic integrity of the generated text. Still, the task at hand is not to evaluate the performance of the language model whereas to evaluate if the generated text has semantic closeness with the context provided. Due to the nature of problem at hand, I have used the following evaluation technique using word embeddings and cosine similarity measures to compute the similarity of the generated text.

\subsection{Cosine similarity measure}
The similarity is carried out between the context word and a list of selected words based on part-of-speech(POS) from the generated sentence. This evaluation is based on the hypothesis that the nouns in the sentence conveys most of the semantic context compared to the rest of the words in the sentence. Thus, only the nouns(NOUN, PROPN) are processed and the cosine similarity of each nouns with the context word is calculated. The average of all the cosine distances in a sentence gives the similarity of the sentence with the context word. Given a sentence $S$ containing nouns $n_i$, the cosine similarity measure is expressed as in equation(\ref{eq:eval})

\begin{equation}
\frac{1}{N}\textstyle \sum_{n_i}^{n_n} m_i.\cos(n_i ,c)
\label{eq:eval}
\end{equation}

\begin{figure}[h]
\centering
\includegraphics[scale=0.27]{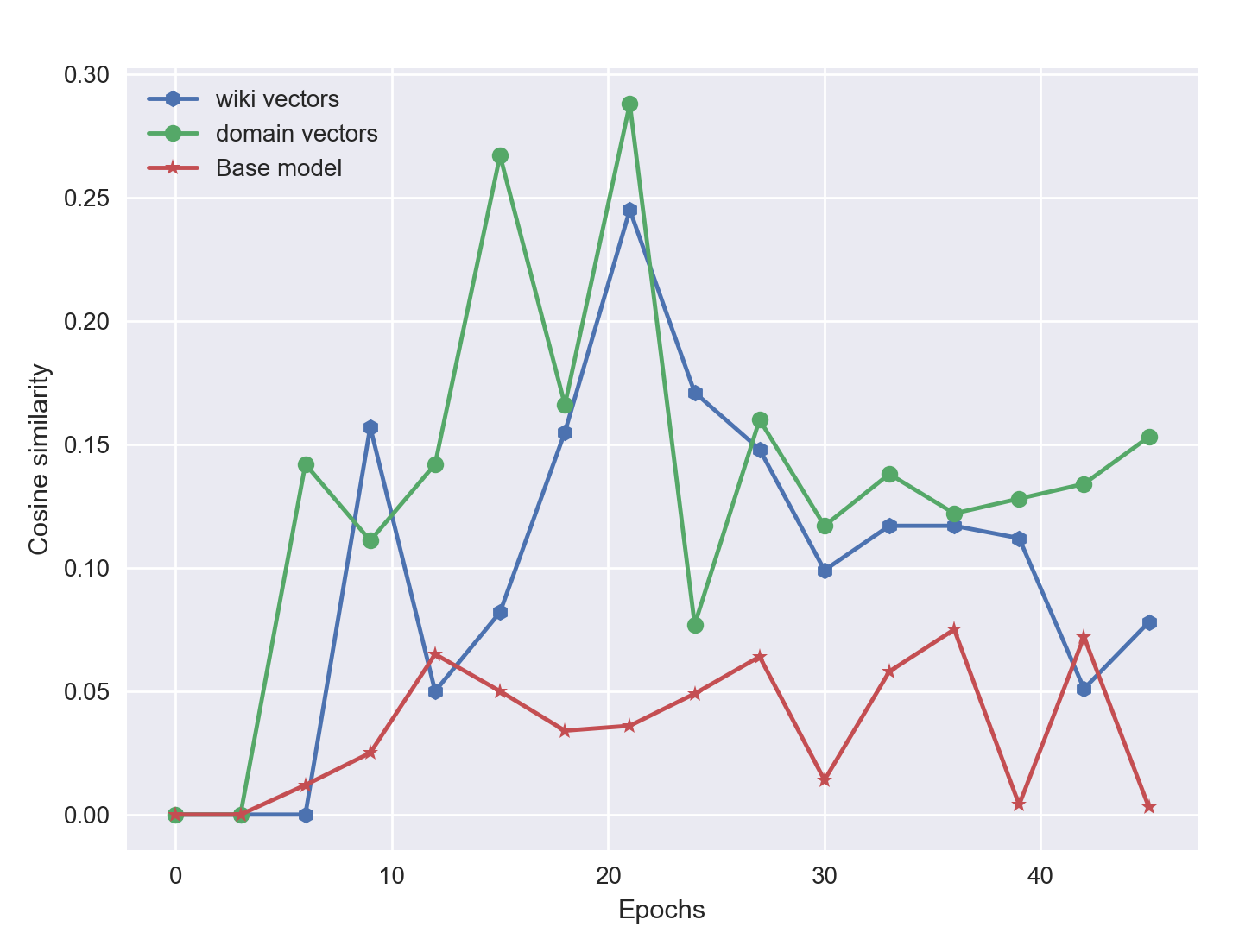}
\caption{Cosine similarity of the generated text with context over training time}
\label{fig:eval}
\end{figure}


The evaluation measure equation(\ref{eq:eval}) is applied on every 3 epochs to both \emph{wiki and domain space} models to measure how good is the semantic closeness between the context provided and the generated text. Figure~\ref{fig:eval} shows a plot of semantic similarity measure of the output over time.

Apart from the context based methods, it is also essential to evaluate them against the base model. The base model performed text generation task without any context vectors. Thus, the usage of the context vectors did not affect the base model at all. The spikes produced by the base model are purely random. It is obvious from the plot that training language models along with context information produce much better results than the base model.

\begin{figure}[h]
\centering
\includegraphics[scale=0.3]{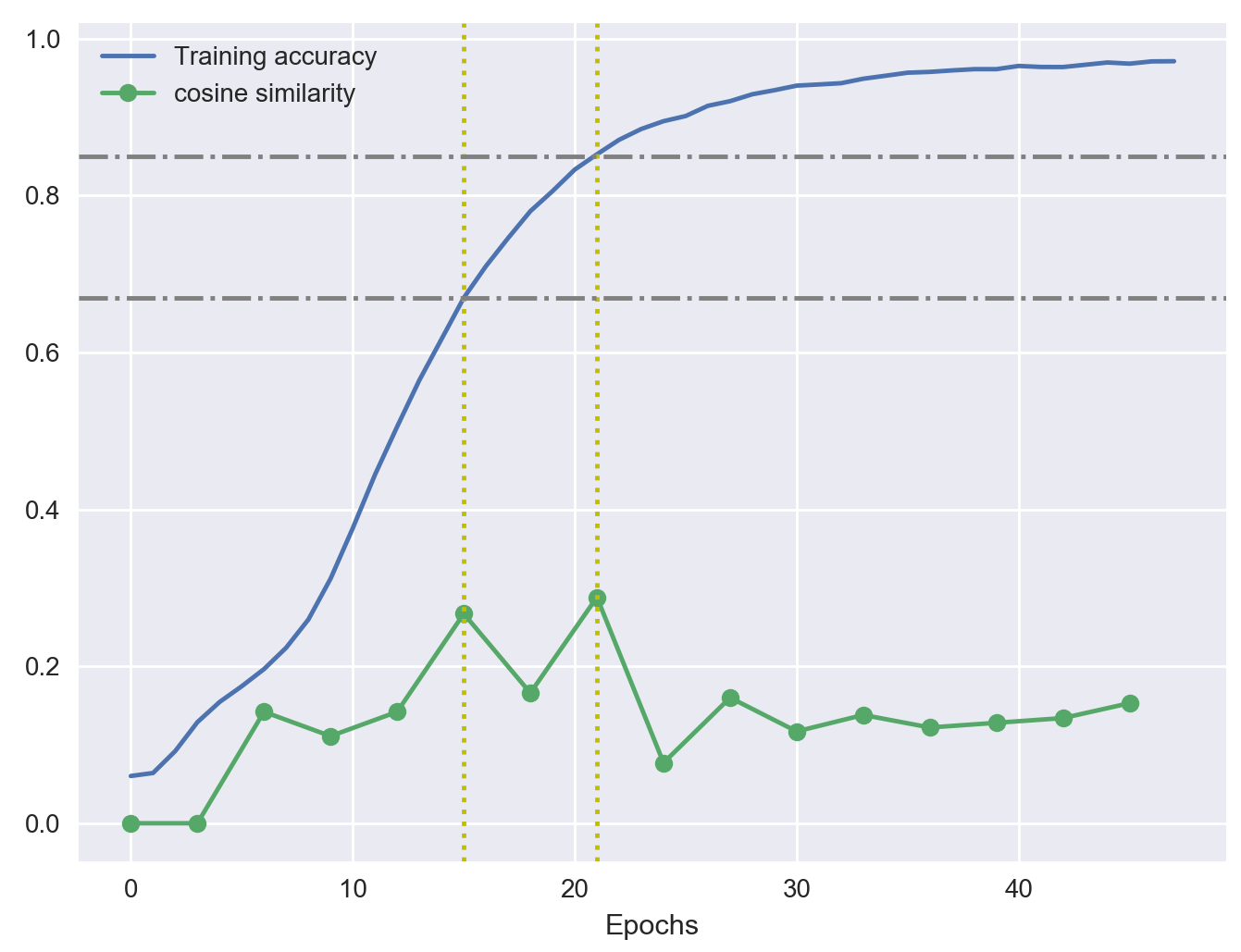}
\caption{Optimum training period obtained from evaluation}
\label{fig:optimum}
\end{figure}

From the plot, it is clear that the context based models' performance increased over time and at a specific point it starts to decrease. One important thing to note is that, cosine similarity measure could go negative, if the words are highly dissimilar. When considering the cosine similarity of the whole generated text, if in case there are few negative values they will be negligible when computing the whole average. By comparison, the model trained on context extraction from the \emph{domain vector space} performed better than the \emph{wiki vector space} embeddings. Around 20 epochs of training, both the models achieved maximum performance. Thus, on plotting the maximum peaks of the domain training model for the given dataset, one could figure out the optimum performance of the model with relative to context based generation. This is shown in the Figure~\ref{fig:optimum}.

From Figure~\ref{fig:optimum}, we could see that a training accuracy between 67\% and 85\% provides a generalizable model for the considered use case.

\section{Discussion}
The paper discussed about the need for applying contextual information to train language models for text generation tasks. LSTM networks have been chosen as the language model for the considered use-case. The main focus of the work is on finding out the best way of extracting context from the sentences, so as to train a text generation model along with the contexts aiming for a better language model. Multiple methods have been attempted for context extraction and among those methods, contexts extracted from word clusters in word vector spaces worked better. Also, it has been clear from the results that the word embeddings generated from the same dataset provided better context vectors. The evaluation method used cosine similarity measures to calculate the semantic closeness between the generated sentences and the provided context. The evaluation results also emphasized that context based models perform better compared to base models. The evaluation method also provided a feedback to the training process to know when the model overfits so as to stop training the model. The proposed system could find potential applications in the following areas:
\begin{itemize}
\item Question answering systems, where instead of following a template based approach the model could identify the context from the question and respond meaningfully concerning the context
\item By keeping track of topics in chat-bots, conversations could be generated based on the previously conversed topics/contexts
\item Besides context vectors defining semantic meanings of sentences, it could rather be used to specify a structure. for instance, when text-generation is trained with context vectors of introduction-content-summary-moral structures. Thus a generated report/story instead of having a random format, a defined form could be expected of the result
\end{itemize}

\subsection{Future work}
\begin{itemize}
\item Attention based models like Transformers can be used as language models instead of memory based networks
\item Other clustering methods like spectral clustering, DBSCAN could be tried out for context extraction
\item More complex evaluation methods using weighted similarities based on POS tags can be considered
\end{itemize}

\bibliographystyle{unsrt}  
\bibliography{literature}  


\end{document}